\documentclass[runningheads]{llncs}

\usepackage[T1]{fontenc}
\usepackage{graphicx}
\usepackage{amsmath}
\usepackage{amssymb}
\usepackage{array}
\usepackage{booktabs}
\usepackage{multirow}
\usepackage{placeins}
\usepackage{url}
\graphicspath{{figures/}}

\begin{document}

\title{FedCVESA: Taking Away Training Data in Federated Learning via Correlation Value Encoding and Segmented Aggregation}
\titlerunning{FedCVESA for TATD in FL}

\author{Chongkai Li\inst{1} \and
Bang Zhang\inst{1} \and
Wenjian Luo\inst{1}\thanks{Corresponding author.}}

\authorrunning{C. Li et al.}

\institute{Guangdong Provincial Key Laboratory of Novel Security Intelligence Technologies,
Institute of Cyberspace Security, School of Computer Science and Technology,
Harbin Institute of Technology, Shenzhen 518055, China}

\maketitle

\begin{abstract}
Federated learning (FL) avoids explicit data exposure by keeping raw data on local clients,
yet privacy risks remain in the training process and the learned model itself.
Recently, centralized Taking Away Training Data (TATD) attacks have shown that malicious training could abuse the memorization capacity of deep models to store and later recover training data.
However, this memorization-based threat has not been systematically studied under FL environments, where multi-client averaging could overwrite encoded training data.
In this paper, we study a white-box TATD attack in which a malicious server selects $n$ target clients from $K$ participating clients and actively writes private training data into the global model during federated training.
We propose FedCVESA, a federated variant of Correlation Value Encoding Attack (CVEA), by adding a Pearson-correlation regularizer to the loss function of target clients,
so that private training data are gradually encoded into selected model parameters, referred to as carrier parameters.
To reduce the overwriting of carrier parameters during server aggregation,
we further propose segmented aggregation over dispersed carrier parameters, preserving selected carrier parameters while keeping standard averaging on the remaining parameters.
Experiments on MNIST, Fashion-MNIST, and CIFAR-10 under Dirichlet non-IID partitions show that the proposed method can steal semantically meaningful private training images from the trained model while maintaining acceptable main-task utility in a controlled proof-of-concept setting.
These results demonstrate that FL can become a parameter-level memorization channel for active TATD attack under the studied white-box malicious-server setting.

\end{abstract}

\section{Introduction}

Federated learning (FL) allows multiple clients to jointly train a model without directly sharing raw training data,
and has therefore become a standard paradigm for privacy-sensitive applications such as healthcare, finance, and mobile intelligence \cite{mcmahan2017fedavg,kairouz2021flopen,yang2019fml,li2020flfuture}.
However, the slogan ``data never leaves the device'' should not be mistaken for end-to-end privacy protection.
Model updates still expose rich information about local training data,
and a sufficiently capable adversary can exploit gradients or parameters to infer private data \cite{zhu2019dlg,geiping2020inverting,nasr2019privacy,wei2020clientprivacy,boenisch2023curious}.

Most existing white-box data leakage attacks against FL focus on gradient inversion.
Representative methods such as DLG, iDLG, and GradInversion recover input training data by matching dummy gradients to observed gradients \cite{zhu2019dlg,zhao2020idlg,yin2021gradinversion,jeon2021gradient}.
These attacks demonstrate that gradients leak substantial information,
yet they remain fundamentally passive: the attacker can only reconstruct whatever information is already exposed by the observed updates.

Centralized Taking Away Training Data (TATD) attacks study a different privacy risk: malicious training can force a model to remember training data.
Early work showed that models can be intentionally trained to remember too much \cite{song2017remember}.
Subsequent attacks encoded training data through parameter combinations, exploited unused model capacity for training-data leakage in a black-box setting, hiding leakage behind backdoor-style triggers, or stored information in output confidences \cite{luo2022pcea,luo2024capacity,yang2022misclassification,yang2023imperceptible,naseem2026confidences}.
These studies reveal that model memorization can be turned into a data-stealing channel in centralized settings.

To the best of our knowledge, existing FL privacy studies have not systematically examined whether such model memorization can be actively exploited to take away target-client training data under federated aggregation.
In FL, training data owned by target clients are mixed with updates from many non-target clients, and standard aggregation could overwrite the carrier parameters before the server can recover the data.
Thus, an FL Taking Away Training Data attack must coordinate local encoding, aggregation-time preservation, and server-side recovery.

We address this gap by studying a white-box TATD attack in FL.
We actively modify the local optimization objective on the $n$ target clients so that private training data are gradually written into a subset of model parameters throughout federated training.
Our method, called FedCVESA, builds on Correlation Value Encoding Attack (CVEA) \cite{song2017remember}
and adapts it to FL with a malicious-server segmented aggregation scheme that reduces overwriting of the encoded training data.

The main contributions of this paper are as follows.
\begin{itemize}
\item We formulate a white-box TATD threat for federated learning, in which $n$ target clients selected from $K$ participating clients are subjected to active memorization-based data encoding during training and a malicious server performs post-aggregation extraction and recovery from the aggregated global model.
\item We propose FedCVESA, which combines local correlation value encoding with segmented aggregation over dispersed carrier parameters, so that training data can be recorded in model parameters while the model remains on the normal FL optimization path.
\item We validate FedCVESA on MNIST, Fashion-MNIST, and CIFAR-10 under Dirichlet non-IID data partitions. The results show that the method can steal semantically meaningful private training images while maintaining main-task utility, highlighting an effectiveness--utility trade-off in this proof-of-concept setting.
\end{itemize}

The source code of FedCVESA is publicly available at \url{https://github.com/MiLab-HITSZ/2026LiFedCVESA}.

\section{Related Work}

\subsection{Data Leakage Attacks in Federated Learning}

DLG~\cite{zhu2019dlg}, iDLG~\cite{zhao2020idlg}, GradInversion~\cite{yin2021gradinversion}, and Inverting Gradients~\cite{geiping2020inverting} recover input training data by matching or regularizing observed gradients.
Recent work extends this risk beyond standard gradients: generative-prior inversion~\cite{jeon2021gradient}, modified-model attacks~\cite{fowl2022robbing}, gradient magnification~\cite{wen2022fishing}, and dishonest-server variants~\cite{boenisch2023curious} improve extraction in harder settings.
SRATTA~\cite{marchand2023sratta} studies recovery under secure aggregation~\cite{bonawitz2017secureagg}, while Cocktail Party Attack~\cite{kariyappa2023cocktail} and TabLeak~\cite{vero2023tableak} study recovery from aggregated or non-image FL signals.
These methods remain mostly passive in the sense that they exploit information already exposed by updates or aggregated models.
Earlier collaborative-learning leakage studies also show that shared training signals can reveal sensitive information beyond labels or aggregate accuracy, including unintended feature leakage and generative reconstruction risks~\cite{hitaj2017ganleakage,melis2019featureleakage}.

Active FL security work studies model poisoning and backdoor attacks that manipulate local training or aggregation behavior~\cite{bhagoji2019adversarial,bagdasaryan2020backdoor,xie2020dba,sun2019backdoor,wang2020attacktails,fang2020localpoisoning,zhang2022neurotoxin}.
These attacks show that adversarial control can change the information carried by federated parameters; FedCVESA uses such control as a memorization channel for recovering training data.

\subsection{Taking Away Training Data Attacks}

Taking Away Training Data (TATD) attacks study how malicious training can make a learned model store and later reveal training data while keeping the main task usable.
This threat is different from membership inference or model inversion \cite{shokri2017membership,fredrikson2015modelinversion}: the attacker targets concrete training data stored in the trained model, beyond the existence of a data point, attribute, or class representative.
The idea can be traced back to Song \textit{et al.} \cite{song2017remember}, who first showed that machine learning models can be intentionally trained to remember too much through parameter-level and label-based encoding channels.
Unintended memorization and training-data extraction have also been studied in broader learning systems, showing that learned models can expose rare or sensitive training sequences under suitable access conditions~\cite{carlini2019secretsharer,carlini2021extracting}.
More recently, Naseem \textit{et al.} \cite{naseem2026confidences} explicitly framed this memorization-based data-theft threat as Taking Away Training Data and proposed Confidence Memory Attack (CMA), which encodes private data into model output confidences.

According to the attacker's access to the trained model, existing TATD attacks can be grouped into three types, i.e., white-box, gray-box, and black-box settings.
In white-box settings, the attacker can inspect model parameters and use them as a storage channel.
Song \textit{et al.} \cite{song2017remember} proposed parameter-level encoding, and Luo \textit{et al.} \cite{luo2022pcea} proposed PCEA, which encodes training data through parameter combinations in a white-box setting.
In gray-box settings, CMA uses output confidences as the encoding and recovery channel \cite{naseem2026confidences}.
In black-box settings, the attacker lacks direct access to model parameters and output confidences, and relies on model outputs.
The Capacity Abuse Attack and its improved version exploit model capacity under black-box access by using malicious data points and label encodings to recover training data after model training \cite{luo2024capacity}.
Other studies improve the concealment of black-box data-leakage attacks by replacing obviously synthetic malicious data points with backdoor-triggered or imperceptibly modified data points \cite{yang2022misclassification,yang2023imperceptible}.

However, TATD attacks are studied mainly in centralized settings, where training is not subject to federated aggregation.
They do not address the FL settings, where encoded training data from target clients can be overwritten by other clients' updates during aggregation.
This leaves open whether active TATD can be made viable in federated learning, where the attacker must coordinate local encoding with server-side aggregation and recovery.

\subsection{CVEA and Parameter-Level Encoding}
\label{subsec:cvea}

Correlation Value Encoding Attack (CVEA), proposed by Song \textit{et al.} \cite{song2017remember}, is the direct technical foundation of FedCVESA.
It turns model parameters into a storage channel by maximizing the statistical correlation between selected carrier parameters and an attacker-chosen signal.
CVEA actively writes information into the model through an encoding objective, giving it a different mechanism from gradient inversion.

CVEA is suitable for our FL setting because its differentiable correlation objective can be inserted into the local client loss and its strength can be controlled to study the stealing quality and utility trade-off.
FedCVESA transfers this parameter-level encoding idea to federated learning, where segmented aggregation is needed to prevent encoded training data from being overwritten by multi-client averaging.

\section{Threat Model}

We consider a synchronous cross-device FL system with one server and $K$ participating clients training a neural network $f(\cdot;w)$.
Here, $w$ denotes the full trainable parameter vector of the neural network, and $w_t$ denotes the global parameter vector at communication round $t$.
In each round, the server broadcasts $w_t$ to these $K$ participating clients.
Each participating client performs local optimization on its private dataset and uploads an updated local parameter vector $w_t^k$.
The server then aggregates these updates to obtain $w_{t+1}$.

A malicious server selects $n$ clients from the $K$ participating clients as targets. We refer to these victims as target clients, and to the remaining $K-n$ clients as non-target clients.
The malicious server knows the model architecture, training pipeline, and has access to model parameters throughout training.
On target clients, the local training objective is manipulated by the attack routine, while the malicious server can customize aggregation on carrier parameters before publishing the next global model.
After aggregation, the malicious server can inspect the global model and run a recovery procedure.
The encoded training data are drawn from private training data held locally by the target clients during normal federated learning.
Under the standard FL data-isolation assumption, these raw training data are not directly accessible to the server.

Overall, the attack goal is to encode private training data during training and later recover them from the global model while preserving acceptable main-task behavior.

\begin{figure}[t]
  \centering
  \includegraphics[width=0.92\textwidth]{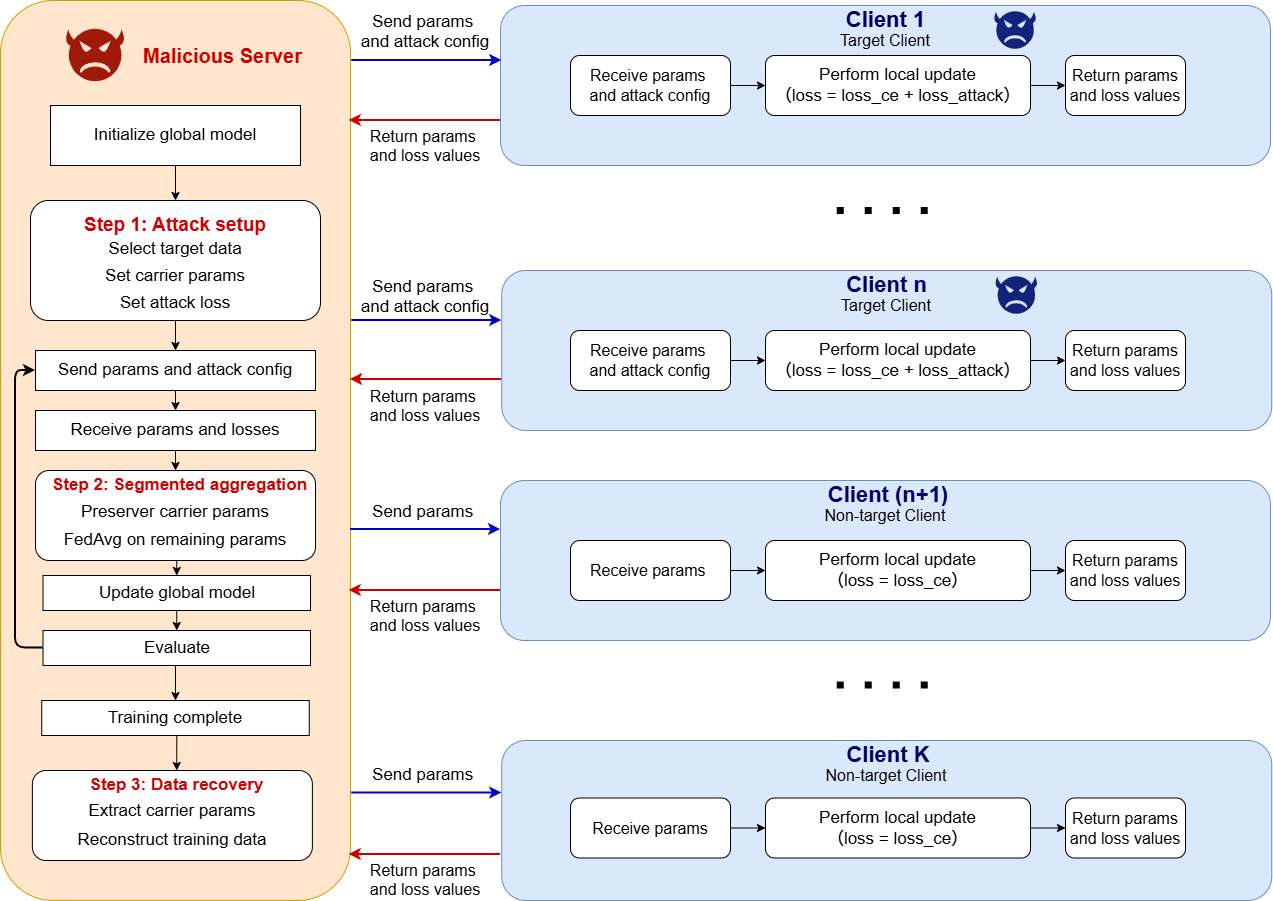}
  \caption{Overall pipeline of the proposed white-box TATD attack in federated learning under a malicious server targeting $n$ clients selected from $K$ participating clients.}
  \label{fig:framework}
\end{figure}

\section{Proposed Method}

In FedCVESA, the malicious server mainly conducts three steps, as illustrated in Fig.~\ref{fig:framework}.
First, the malicious server configures the attack by specifying the target clients, the target-sample indices or selection rule on those clients, the carrier parameters, and the attack loss; on each target client, this attack loss is added as a regularization term to the normal local training objective so that the selected private training data are gradually written into carrier parameters.
Second, the malicious server applies segmented aggregation to prevent the encoded training data from being overwritten by federated averaging~\cite{mcmahan2017fedavg} while leaving the remaining parameters on the standard FL update path.
Third, the malicious server extracts the preserved values from the global model and rearranges them to recover the training data.
The following subsections introduce correlation value encoding, segmented aggregation over dispersed carrier parameters, and server-side extraction and recovery.
The clients in FedCVESA are divided into two groups: target clients selected by the malicious server and non-target clients, i.e., the remaining participating clients.

\subsection{Correlation Value Encoding}

Standard client training minimizes a classification loss $L_{\text{ce}}$.
To turn local training into a parameter-encoding process,
we augment the local objective on target clients with a correlation regularizer:
\begin{equation}
L_{\text{total}} = L_{\text{ce}} - \gamma L_{\text{corr}},
\label{eq:total-loss}
\end{equation}
where $\gamma$ controls attack strength. Following the CVEA principle introduced in Section~\ref{subsec:cvea},
$L_{\text{corr}}$ is the federated instantiation of the malicious correlation term and encourages the model parameters to become statistically aligned with private training data.

In our implementation, the encoded training data are derived from private training images on the target clients after unified preprocessing.
Let $\theta$ denote the flattened carrier-parameter vector used for attack carrying,
and let $d$ denote the flattened to-be-stolen training-data vector.
Both vectors have the same length, and the subscript $i$ indexes their corresponding elements.
The terms $\bar{\theta}$ and $\bar{d}$ denote the empirical means of $\theta$ and $d$, respectively.
We instantiate the CVEA-style malicious term with Pearson-style correlation~\cite{song2017remember}:
\begin{equation}
L_{\text{corr}} = \frac{\sum_i(\theta_i-\bar{\theta})(d_i-\bar{d})}
{\sqrt{\sum_i(\theta_i-\bar{\theta})^2}\sqrt{\sum_i(d_i-\bar{d})^2}}.
\label{eq:corr}
\end{equation}
Maximizing this term gradually biases the optimization trajectory so that training data are written into the parameter space.
Smaller $\gamma$ better preserves main-task utility but yields weaker encoding; larger $\gamma$ improves Stealing quality but risks disrupting the main task.

\subsection{Segmented Aggregation over Dispersed Carrier Parameters}

The attack requires a sufficient number of carrier parameters to record training data.
A naive design is to use a contiguous parameter block.
However, when too many carrier parameters are selected at contiguous positions, they may disproportionately perturb one or more network layers, and then could distort that layer's normal representation and noticeably degrade the main task.
We therefore flatten the full model and construct a set of dispersed carrier-parameter indices across the entire parameter vector:
\begin{equation}
s_r = \left\lfloor \frac{r(L-1)}{T-1} \right\rfloor,\quad r=0,1,\dots,T-1,
\label{eq:dispersed-index}
\end{equation}
where $L$ is the flattened parameter vector length and $T$ is the total number of carrier parameters.
These indices are then logically partitioned into client-specific subsets.
This design does not increase the encoding budget; it disperses a fixed number of carrier parameters across the model, preventing the encoding burden from being concentrated in any single layer.

During server aggregation, let $\mathcal{K}$ denote the set of $K$ participating clients, and let $\mathcal{C}_{\text{target}}\subseteq\mathcal{K}$ denote the $n$ target clients among them.
For each target client $c\in\mathcal{C}_{\text{target}}$, $\Omega_c$ denotes the carrier-parameter index set assigned to that client.
These client-specific sets are disjoint, and their union is the full carrier-parameter index set:
\begin{equation}
\Omega_{\text{attack}}
=
\bigcup_{c\in\mathcal{C}_{\text{target}}}\Omega_c,
\quad
\Omega_c\cap\Omega_{c'}=\varnothing\ (c\neq c').
\label{eq:attack-index-set}
\end{equation}
Here, $c$ and $c'$ denote two different target clients in $\mathcal{C}_{\text{target}}$.
The non-carrier parameter indices are the complement of $\Omega_{\text{attack}}$ in the full flattened parameter-index set, and the corresponding parameters still follow standard averaging.
For each $j\in\Omega_{\text{attack}}$, let $c$ denote the unique target client satisfying $j\in\Omega_c$.
For carrier parameters, the malicious server keeps the value uploaded by the target client assigned to the corresponding carrier-parameter index; for all remaining parameters, it applies standard averaging:
\begin{equation}
w_{t+1}(j)=
\begin{cases}
w_t^c(j), & j\in\Omega_{\text{attack}}\\[4pt]
\frac{1}{|\mathcal{K}|}\sum_{k\in\mathcal{K}} w_t^k(j), & j\notin\Omega_{\text{attack}}.
\end{cases}
\label{eq:seg-agg}
\end{equation}
Here, $w_t^c(j)$ denotes the uploaded value from the target client assigned to carrier-parameter index $j$.
This segmented aggregation preserves selected carrier parameters while keeping the model on the normal FL optimization path.
Together with dispersed carrier-parameter placement, it mitigates both aggregation-induced overwriting of the encoded training data and local disruption caused by concentrating too many carrier parameters in one layer.

\begin{figure}[t]
  \centering
  \includegraphics[width=0.92\textwidth]{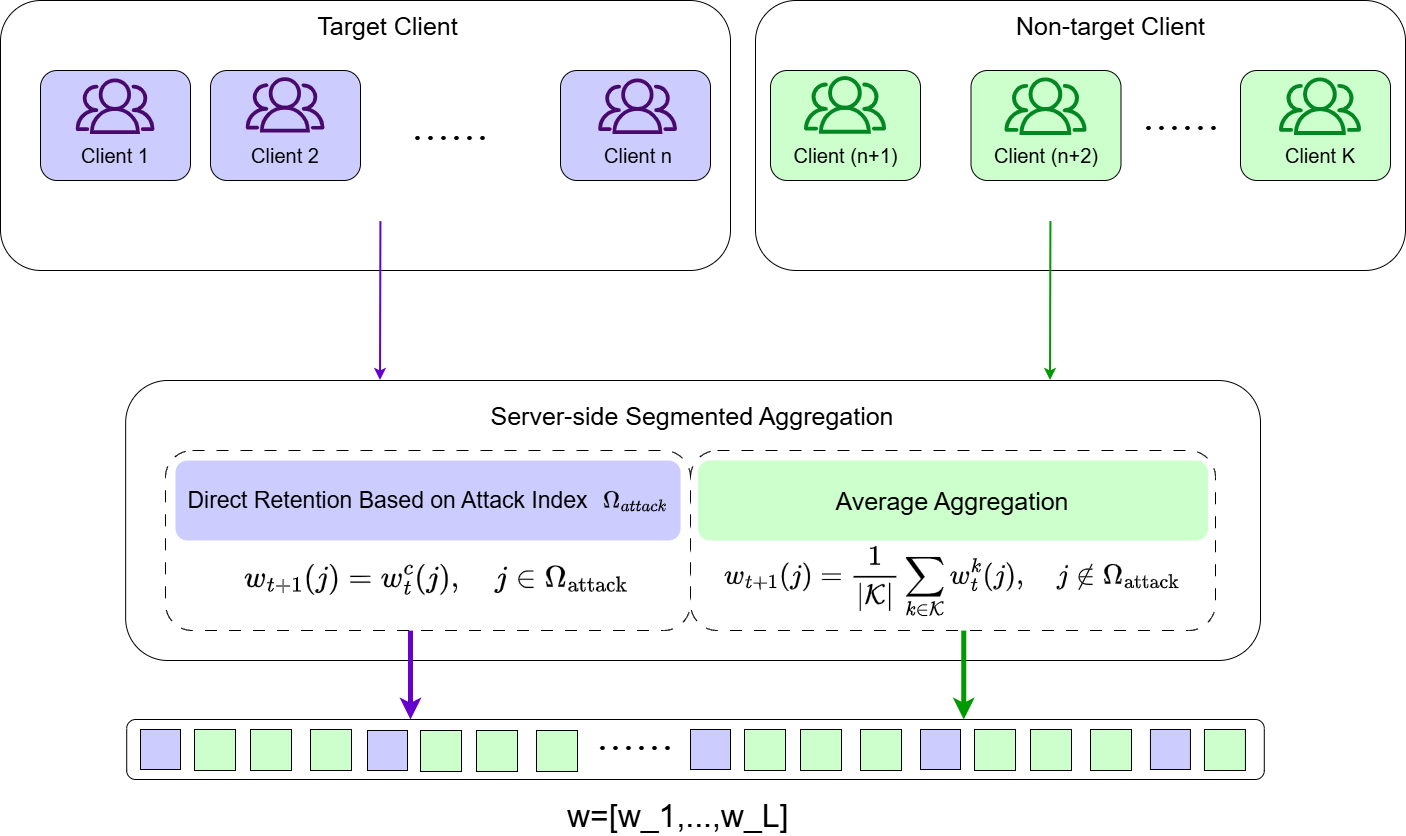}
  \caption{Segmented aggregation with dispersed carrier parameters. Logical segments are assigned to target clients, but the carrier parameters are globally scattered in the parameter vector.}
  \label{fig:segagg}
\end{figure}

\subsection{Server-Side Extraction and Recovery}

After training, the malicious server extracts the preserved carrier parameters from the global model and decodes them using the same index order as in local encoding and segmented aggregation.
Using the previously defined client-specific carrier-parameter index set $\Omega_c$, the server extracts the encoded vector~\cite{song2017remember}:
\begin{equation}
z_c = [w_{t+1}(j)]_{j\in\Omega_c},\quad c\in\mathcal{C}_{\text{target}} .
\label{eq:index-extract}
\end{equation}
For a single training image with preprocessed shape $C\times H\times W$, we set $|\Omega_c|=S=CHW$ and recover the image by min--max normalization~\cite{han2011datamining} and reshaping:
\begin{equation}
\begin{aligned}
\hat{x}_c &= \operatorname{Reshape}(\operatorname{Norm}(z_c), C, H, W),\\
\operatorname{Norm}(z_c)_i &=
\frac{z_{c,i}-\min(z_c)}{\max(z_c)-\min(z_c)+\epsilon}.
\end{aligned}
\label{eq:recover-normalize}
\end{equation}
Here, $\operatorname{Norm}(\cdot)$ maps extracted parameter values to the image value range, and $\epsilon=10^{-8}$ avoids division by zero.
When a target client carries $q$ training images, we set $|\Omega_c|=qCHW$, split $z_c$ into $q$ consecutive vectors of length $CHW$, and apply the same normalization and reshaping procedure to each vector.
This recovery step does not require an additional optimization process; it depends on using the same index rule, encoding length, and reshaping convention across local encoding, segmented aggregation, and server-side extraction.
The resulting objective is to obtain semantically meaningful private training images under the studied FL aggregation setting.

\section{Experiments}

\subsection{Experimental Setup}

We implement FedCVESA in a PyTorch-based federated learning simulator \cite{paszke2019pytorch}.
The experiments are conducted on MNIST, Fashion-MNIST, and CIFAR-10 \cite{lecun1998mnist,xiao2017fashionmnist,krizhevsky2009cifar10}.
Each dataset is partitioned across clients using Dirichlet label-skew sampling with concentration parameter $\alpha=0.5$ \cite{hsu2019measuring}; in our implementation, we enforce a minimum of 100 training samples per client.
Specifically, for each class label $\ell$, we sample a client-allocation vector from a symmetric Dirichlet distribution:
\begin{equation}
\boldsymbol{\pi}_{\ell}
=
(\pi_{\ell,1},\pi_{\ell,2},\dots,\pi_{\ell,K})
\sim
\operatorname{Dir}_K(\alpha,\alpha,\dots,\alpha),
\quad
\sum_{k=1}^{K}\pi_{\ell,k}=1 .
\label{eq:dirichlet-partition}
\end{equation}
The value $\pi_{\ell,k}$ determines the fraction of samples with label $\ell$ assigned to client $k$.
A smaller $\alpha$ produces more heterogeneous client label distributions, and $\alpha=0.5$ is used to simulate the non-IID FL setting in our experiments.
MNIST and Fashion-MNIST are evaluated as normalized grayscale inputs, while CIFAR-10 uses normalized RGB inputs at $32\times32$ resolution.

We use $K=10$ participating clients, 100 global communication rounds, segmented aggregation, and dispersed carrier-parameter placement.
The default setting selects $n=5$ target clients from the $K$ participating clients and uses one private training image per target client.
We use a CNN for MNIST and Fashion-MNIST and ResNet18~\cite{he2016deep} for CIFAR-10.
Table~\ref{tab:model-architectures} summarizes the main components, activation choices, and parameter count of each model.

\begin{table}[t]
\centering
\caption{Model configurations used in the experiments.}
\label{tab:model-architectures}
\small
\setlength{\tabcolsep}{4pt}
\begin{tabular}{@{}llllr@{}}
\toprule
Dataset & Model & Components & Act. & Params \\
\midrule
MNIST & CNN & 2 conv + 3 FC & LReLU, ReLU & 3.37M \\
Fashion-MNIST & CNN & 2 conv + 3 FC & LReLU, ReLU & 3.37M \\
CIFAR-10 & ResNet18 & CIFAR ResNet18 & SiLU & 11.17M \\
\bottomrule
\end{tabular}
\end{table}

All models use GroupNorm.
The CNN uses average pooling after each convolutional block and dropout after the first fully connected layer.
The CIFAR-adapted ResNet18 uses a $3\times3$ stem, no initial max-pooling, adaptive average pooling, and the standard ResNet18 block configuration $[2,2,2,2]$.

For MNIST and Fashion-MNIST, the local learning rate is 0.01, the number of local epochs is 10, and the local batch size is 16.
For CIFAR-10, the local learning rate is 0.03, the number of local epochs is 1, and the local batch size is 64.

We evaluate FedCVESA from both main-task utility and Stealing quality perspectives.
Main-task utility is primarily measured by final test accuracy at the final global round.
Stealing quality is measured by mean absolute percentage error (MAPE) between reconstructed training data and the corresponding private training data; lower MAPE indicates better Stealing quality.
For training data $x$ and its reconstruction $\hat{x}$ with $P$ pixels, we compute:
\begin{equation}
\operatorname{MAPE}(x,\hat{x})
=
\frac{1}{P}\sum_{i=1}^{P}
\frac{|x_i-\hat{x}_i|}{|x_i|+\epsilon},
\label{eq:mape}
\end{equation}
where $\epsilon=10^{-8}$ is a small constant used to avoid division by zero for near-black pixels.
For the no-attack baseline ($\gamma=0$), MAPE is not reported because no training data are encoded.

The reported experiments focus on three questions.
First, we vary the attack strength over $\gamma\in\{0,0.05,0.2,0.5,1.0\}$ with $n=5$ target clients, one training image per client, segmented aggregation, and dispersed placement.
Second, we vary the number of target clients $n$ over $\{1,2,3,4,5,10\}$ while using $\gamma=0.5$, one training image per client, segmented aggregation, and dispersed placement.
Third, we ablate the carrier-parameter placement strategy by comparing contiguous placement with dispersed placement under $n=10$ target clients and $q\in\{10,50\}$ training images per target client.

\subsection{Attack Strength Analysis}

\begin{figure}[t]
  \centering
  \includegraphics[width=\textwidth]{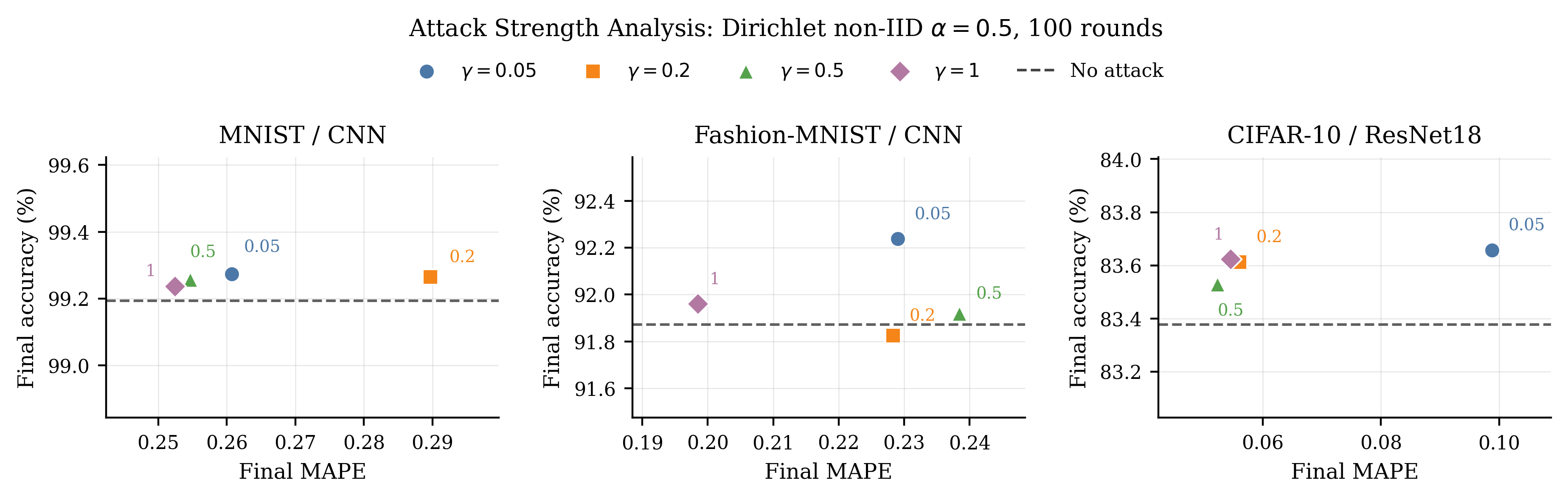}
  \caption{Attack-strength scan under Dirichlet non-IID partitions with five target clients and one training image per client. Main-task utility is shown by final accuracy, with the no-attack baseline indicated by the dashed line, while stealing quality is measured by MAPE.}
  \label{fig:gamma-tradeoff}
\end{figure}

We evaluate the effect of attack strength by varying $\gamma$ with $K=10$ participating clients, $n=5$ target clients, one training image per target client, segmented aggregation, and dispersed placement.
Fig.~\ref{fig:gamma-tradeoff} summarizes the final main-task utility--stealing quality trade-off by plotting final accuracy together with the no-attack baseline and MAPE at the final global round.

\begin{figure}[!t]
  \centering
  \includegraphics[width=\textwidth]{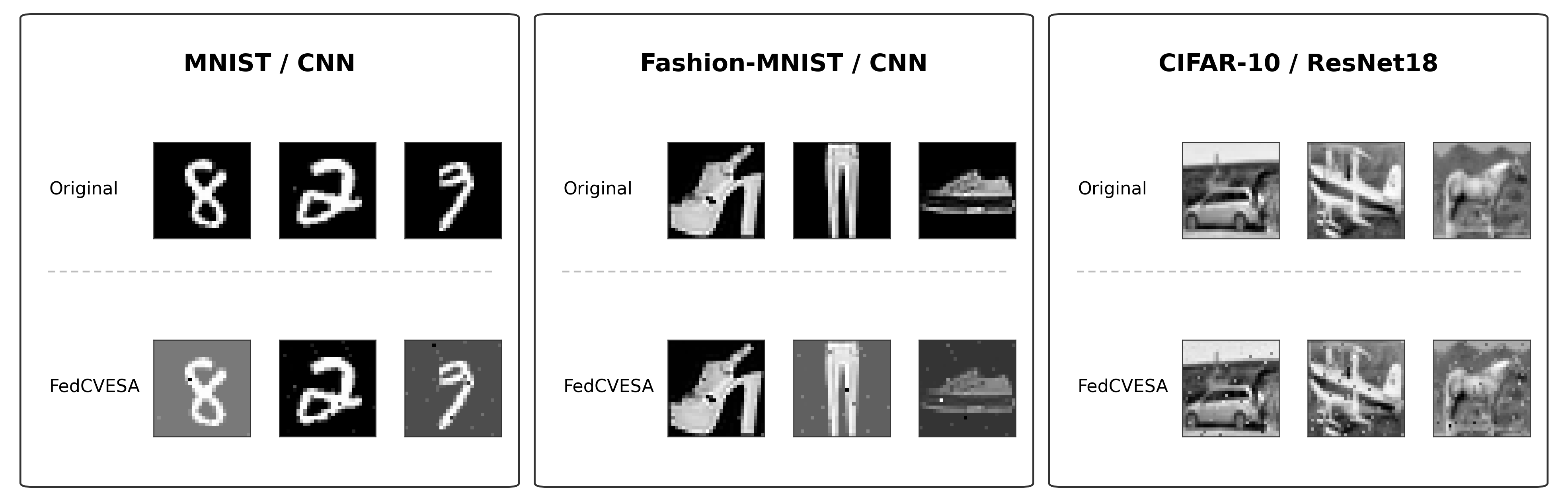}
  \caption{Qualitative recovery examples under the representative Dirichlet non-IID setting with $\gamma=0.5$, $n=5$ target clients, one training image per target client, segmented aggregation, and dispersed placement. For each dataset, the first row shows original private training images and the second row shows the corresponding images recovered from the trained global model.}
  \label{fig:qualitative-recovery}
\end{figure}

Under Dirichlet non-IID partitions, main-task utility remains stable across attack strengths.
MNIST maintains final accuracy between 99.23\% and 99.30\%, and its lowest final MAPE appears at $\gamma=1.0$ with 0.2524.
Fashion-MNIST keeps final accuracy between 89.99\% and 90.45\%, and its lowest final MAPE is also obtained at $\gamma=1.0$ with 0.1985.
These results show that larger attack strength can improve Stealing quality in some cases, although the final MAPE does not decrease monotonically.
This behavior reflects the role of $\gamma$ as a trade-off coefficient.
Increasing $\gamma$ gives the correlation term more influence over local optimization and can align carrier parameters more strongly with the target image vector.
The final signal is also shaped by stochastic training, non-IID client data, and segmented aggregation, making recovery quality non-monotonic.

CIFAR-10 with ResNet18 also shows a stable main-task utility profile under the attack-strength scan.
Final accuracy ranges from 80.77\% to 81.21\%, and the lowest final MAPE appears at $\gamma=0.5$ with 0.0523.
The CIFAR-10 results suggest that Stealing quality depends on the interaction between attack strength, model architecture, optimization dynamics, and carrier-parameter placement.
The CIFAR-10 case is informative because the payload is larger and the model has a deeper representation hierarchy than the CNN used for MNIST-style datasets.
The best final MAPE at $\gamma=0.5$ suggests that an intermediate attack strength creates a recoverable parameter-level signal while keeping local optimization stable.
The narrow final-accuracy range indicates that segmented aggregation confines most attack influence to carrier positions and leaves the remaining parameters on a normal FedAvg-like path.

To complement the quantitative MAPE results, Fig.~\ref{fig:qualitative-recovery} shows qualitative recovery examples under the representative default attack setting.
The examples illustrate whether the extracted carrier-parameter values preserve visually recognizable training-data semantics after recovery.
The visual comparison is important because MAPE can be sensitive to near-zero pixels and dataset-specific intensity distributions.
For image data, semantic recognizability matters alongside pixel-wise error, and the recovered examples check whether carrier parameters contain meaningful target-sample structure.

\subsection{Target-Client Scale Analysis}

\begin{figure}[!t]
  \centering
  \includegraphics[width=0.90\textwidth]{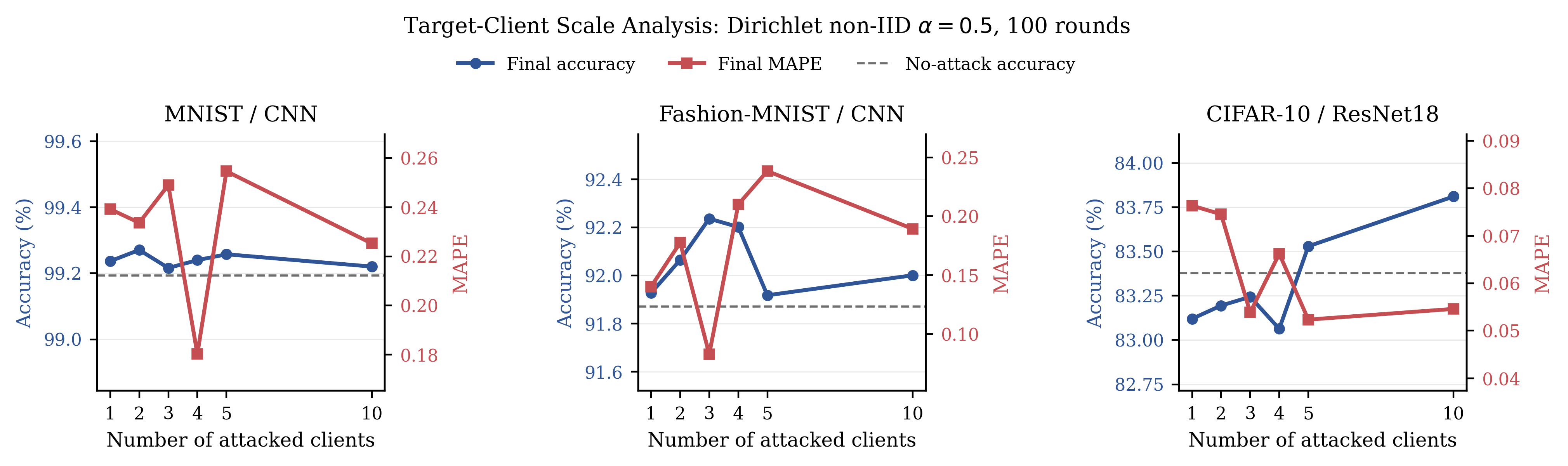}
  \caption{Effect of target-client scale under Dirichlet non-IID partitions with $\gamma=0.5$ and one training image per client. Main-task utility is shown by final accuracy, with the no-attack baseline indicated by the dashed line, while stealing quality is measured by MAPE.}
  \label{fig:num-steal-effect}
\end{figure}

We evaluate the effect of target-client scale by using $K=10$ participating clients, $\gamma=0.5$, one training image per target client, segmented aggregation, and dispersed placement, while varying the number of target clients $n$.
This setting evaluates whether FedCVESA remains stable as the amount of simultaneously stolen private training images increases within the participating-client set.
Fig.~\ref{fig:num-steal-effect} plots final accuracy together with the no-attack baseline and MAPE against the number of target clients $n$.

For MNIST and Fashion-MNIST, varying $n$ has little effect on classification accuracy.
MNIST maintains final accuracy between 99.23\% and 99.32\%, with the lowest final MAPE of 0.1802 at $n=4$.
Fashion-MNIST remains between 89.53\% and 90.26\% final accuracy, with the lowest final MAPE of 0.0827 at $n=3$.
The stability comes from the fixed per-client payload and disjoint carrier subsets.
Increasing $n$ raises the total number of stolen images while keeping each target client's encoding burden unchanged.
Segmented aggregation also prevents non-target updates from averaging away selected carrier values, helping keep accuracy close to the no-attack baseline.

CIFAR-10 with ResNet18 also remains stable in main-task utility, with final accuracy between 80.70\% and 81.59\%.
Its lowest final MAPE appears at $n=5$ with 0.0523.
Overall, increasing the number of target clients does not lead to a monotonic change in MAPE, and the final main-task utility remains stable across the tested target-client scales.
The non-monotonic MAPE trend is also expected because changing $n$ changes target-client identities, non-IID local distributions, selected images, and client-specific carrier subsets.
Recovery quality therefore varies across client-image pairs and is not determined only by the number of targets.
These results show that FedCVESA can scale from a small number of victims to all participating clients in this controlled setting without a visible collapse in the primary classification task.

\subsection{Carrier-Parameter Placement Ablation}

We examine how carrier-parameter placement affects the attack under different payloads.
We use $K=10$ participating clients, set $\gamma=0.5$, and apply segmented aggregation.
We compare two placement strategies: \textit{contiguous placement}, which uses the first $N$ carrier parameters of the flattened model parameter vector, and \textit{dispersed placement}, which uniformly samples $N$ dispersed carrier parameters across the full flattened parameter vector according to Eq.~\ref{eq:dispersed-index}.
The ablation uses $n=10$ target clients and varies the number of encoded training images per target client over $q\in\{10,50\}$.

\begin{table}[t]
  \centering
  \caption{Carrier-parameter placement ablation under Dirichlet non-IID partitions.}
  \label{tab:placement-ablation}
  \small
  \begin{tabular*}{\textwidth}{@{\extracolsep{\fill}}ll l rr@{}}
    \toprule
    Dataset & $q$ & Placement & Final Acc. (\%) & Final MAPE \\
    \midrule
    MNIST & 10 & Contiguous & 99.12 & 0.0147 \\
          &    & Dispersed  & 99.30 & 0.2313 \\
    Fashion-MNIST & 10 & Contiguous & 89.60 & 0.0259 \\
                  &    & Dispersed  & 90.04 & 0.1830 \\
    CIFAR-10 & 10 & Contiguous & 77.25 & 0.1481 \\
             &    & Dispersed  & 81.07 & 0.0937 \\
    \midrule
    MNIST & 50 & Contiguous & 99.09 & 0.0639 \\
          &    & Dispersed  & 99.30 & 0.2225 \\
    Fashion-MNIST & 50 & Contiguous & 89.63 & 0.0508 \\
                  &    & Dispersed  & 90.41 & 0.2072 \\
    CIFAR-10 & 50 & Contiguous & 75.71 & 0.2012 \\
             &    & Dispersed  & 80.85 & 0.1589 \\
    \bottomrule
  \end{tabular*}
\end{table}

Table~\ref{tab:placement-ablation} shows that the placement effect is dataset- and payload-dependent.
Contiguous placement obtains lower final MAPE on MNIST and Fashion-MNIST across the tested payloads, indicating that early contiguous parameters can serve as an efficient storage channel for these simpler datasets.
The main-task utility profile changes as the payload and task complexity increase.
On CIFAR-10, dispersed placement gives lower final MAPE for all tested payloads and substantially higher final accuracy under larger payloads, improving accuracy by +3.82 percentage points at $q=10$ and +5.14 percentage points at $q=50$.
These results support dispersed placement as the more stable default choice for larger-scale stealing and more complex data, especially when maintaining main-task utility is important.
The difference between contiguous and dispersed placement highlights a practical tension in parameter-level encoding.
Contiguous placement creates a compact storage region whose values are easier to align with the target vector, explaining its lower MAPE on MNIST and Fashion-MNIST.
For CIFAR-10, this concentration places heavier encoding pressure on a small part of the model and leads to a clearer utility cost.
Dispersed placement spreads carrier indices across the flattened model, reducing the chance that one functional component is dominated by the memorization objective.
The ablation therefore motivates dispersed placement as the default FedCVESA design for complex data and larger payloads, where utility preservation is important.

\section{Conclusion}

This paper studies a white-box TATD attack in federated learning.
By adapting CVEA to the FL setting, FedCVESA turns local optimization into an active memorization process and uses segmented aggregation over dispersed carrier parameters to reduce overwriting of encoded training data.
Experiments on MNIST, Fashion-MNIST, and CIFAR-10 under Dirichlet non-IID partitions show that the proposed method can steal semantically meaningful private training images while preserving acceptable main-task utility in a range of settings.
These findings provide proof-of-concept evidence indicating that,
under strong white-box adversaries, FL may function as a parameter-level memorization channel.

\begin{credits}
\subsubsection{\ackname}
This study is supported by the National Natural Science Foundation of China (No.~62576121),
the Shenzhen Ocean Economic Innovation and Development Demonstration Project (No.~HYSF\allowbreak2026008),
and the Shenzhen Science and Technology Program (No.~ZDSYS202106\allowbreak230918\allowbreak09029).

\subsubsection{\discintname}
The authors have no competing interests to declare that are relevant to the content of this article.
\end{credits}

\bibliographystyle{splncs04}
\bibliography{iconip_refs}

\begin{thebibliography}{10}
\providecommand{\url}[1]{\texttt{#1}}
\providecommand{\urlprefix}{URL }
\providecommand{\doi}[1]{https://doi.org/#1}

\bibitem{bagdasaryan2020backdoor}
Bagdasaryan, E., Veit, A., Hua, Y., Estrin, D., Shmatikov, V.: How to backdoor federated learning. In: Proceedings of the 23rd International Conference on Artificial Intelligence and Statistics. pp. 2938--2948 (2020)

\bibitem{bhagoji2019adversarial}
Bhagoji, A.N., Chakraborty, S., Mittal, P., Calo, S.: Analyzing federated learning through an adversarial lens. In: Proceedings of the 36th International Conference on Machine Learning. pp. 634--643 (2019)

\bibitem{boenisch2023curious}
Boenisch, F., Dziedzic, A., Schuster, R., Shamsabadi, A., Shumailov, I., Papernot, N.: When the curious abandon honesty: Federated learning is not private. In: 2023 IEEE European Symposium on Security and Privacy. pp. 175--199 (2023)

\bibitem{bonawitz2017secureagg}
Bonawitz, K., Ivanov, V., Kreuter, B., Marcedone, A., McMahan, H.B., Patel, S., Ramage, D., Segal, A., Seth, K.: Practical secure aggregation for privacy-preserving machine learning. In: Proceedings of the 2017 ACM SIGSAC Conference on Computer and Communications Security. pp. 1175--1191 (2017)

\bibitem{carlini2019secretsharer}
Carlini, N., Liu, C., Erlingsson, {\'U}., Kos, J., Song, D.: The secret sharer: Evaluating and testing unintended memorization in neural networks. In: 28th USENIX Security Symposium. pp. 267--284 (2019)

\bibitem{carlini2021extracting}
Carlini, N., Tram{\`e}r, F., Wallace, E., Jagielski, M., Herbert-Voss, A., Lee, K., Roberts, A., Brown, T., Song, D., Erlingsson, {\'U}., Oprea, A., Raffel, C.: Extracting training data from large language models. In: 30th USENIX Security Symposium. pp. 2633--2650 (2021)

\bibitem{fang2020localpoisoning}
Fang, M., Cao, X., Jia, J., Gong, N.Z.: Local model poisoning attacks to {Byzantine}-robust federated learning. In: 29th USENIX Security Symposium. pp. 1605--1622 (2020)

\bibitem{fowl2022robbing}
Fowl, L., Geiping, J., Czaja, W., Goldblum, M., Goldstein, T.: Robbing the {Fed}: Directly obtaining private data in federated learning with modified models. In: International Conference on Learning Representations (2022)

\bibitem{fredrikson2015modelinversion}
Fredrikson, M., Jha, S., Ristenpart, T.: Model inversion attacks that exploit confidence information and basic countermeasures. In: Proceedings of the 22nd ACM SIGSAC Conference on Computer and Communications Security. pp. 1322--1333 (2015). \doi{10.1145/2810103.2813677}

\bibitem{geiping2020inverting}
Geiping, J., Bauermeister, H., Dr{\"o}ge, H., Moeller, M.: Inverting gradients: How easy is it to break privacy in federated learning? In: Advances in Neural Information Processing Systems. vol.~33 (2020)

\bibitem{han2011datamining}
Han, J., Kamber, M., Pei, J.: Data Mining: Concepts and Techniques. Elsevier, 3 edn. (2011)

\bibitem{he2016deep}
He, K., Zhang, X., Ren, S., Sun, J.: Deep residual learning for image recognition. In: Proceedings of the IEEE Conference on Computer Vision and Pattern Recognition. pp. 770--778 (2016)

\bibitem{hitaj2017ganleakage}
Hitaj, B., Ateniese, G., P{\'e}rez-Cruz, F.: Deep models under the {GAN}: Information leakage from collaborative deep learning. In: Proceedings of the 2017 ACM SIGSAC Conference on Computer and Communications Security. pp. 603--618 (2017)

\bibitem{hsu2019measuring}
Hsu, T.M.H., Qi, H., Brown, M.: Measuring the effects of non-identical data distribution for federated visual classification. In: Proceedings of the NeurIPS Workshop on Federated Learning for Data Privacy and Confidentiality (2019)

\bibitem{jeon2021gradient}
Jeon, J., Lee, K., Oh, S., Ok, J.: Gradient inversion with generative image prior. In: Advances in Neural Information Processing Systems. vol.~34, pp. 29898--29908 (2021)

\bibitem{kairouz2021flopen}
Kairouz, P., McMahan, H.B., Avent, B., Bellet, A., Bennis, M., Bhagoji, A.N., Bonawitz, K.A., Charles, Z., Cormode, G., Cummings, R., et~al.: Advances and open problems in federated learning. Foundations and Trends in Machine Learning  \textbf{14}(1--2),  1--210 (2021)

\bibitem{kariyappa2023cocktail}
Kariyappa, S., Guo, C., Maeng, K., Xiong, W., Su, G., Qureshi, M.K., Lee, H.H.S.: {Cocktail Party Attack}: Breaking aggregation-based privacy in federated learning using independent component analysis. In: Proceedings of the 40th International Conference on Machine Learning. pp. 15884--15899 (2023)

\bibitem{krizhevsky2009cifar10}
Krizhevsky, A.: Learning multiple layers of features from tiny images. Tech. rep., University of Toronto (2009)

\bibitem{lecun1998mnist}
LeCun, Y., Bottou, L., Bengio, Y., Haffner, P.: Gradient-based learning applied to document recognition. Proceedings of the IEEE  \textbf{86}(11),  2278--2324 (1998)

\bibitem{li2020flfuture}
Li, T., Sahu, A.K., Talwalkar, A., Smith, V.: Federated learning: Challenges, methods, and future directions. IEEE Signal Processing Magazine  \textbf{37}(3),  50--60 (2020)

\bibitem{luo2022pcea}
Luo, W., Zhang, L., Han, P., Liu, C., Zhuang, R.: Taking away both model and data: Remember training data by parameter combinations. IEEE Transactions on Emerging Topics in Computational Intelligence  \textbf{6}(6),  1427--1437 (2022). \doi{10.1109/TETCI.2022.3182415}

\bibitem{luo2024capacity}
Luo, W., Zhang, L., Wu, Y., Liu, C., Han, P., Zhuang, R.: {Capacity Abuse Attack} of deep learning models without need of label encodings. IEEE Transactions on Artificial Intelligence  \textbf{5}(2),  814--826 (2024)

\bibitem{marchand2023sratta}
Marchand, T., Loeb, R., Marteau-Ferey, U., Ogier Du~Terrail, J., Pignet, A.: {SRATTA}: Sample re-attribution attack of secure aggregation in federated learning. In: Proceedings of the 40th International Conference on Machine Learning. pp. 23886--23914 (2023)

\bibitem{mcmahan2017fedavg}
McMahan, H.B., Moore, E., Ramage, D., Hampson, S., Arcas, B.A.y.: Communication-efficient learning of deep networks from decentralized data. In: Proceedings of the 20th International Conference on Artificial Intelligence and Statistics. pp. 1273--1282 (2017)

\bibitem{melis2019featureleakage}
Melis, L., Song, C., De~Cristofaro, E., Shmatikov, V.: Exploiting unintended feature leakage in collaborative learning. In: 2019 IEEE Symposium on Security and Privacy. pp. 691--706 (2019)

\bibitem{naseem2026confidences}
Naseem, M.L., Li, C., Zhang, B., Luo, W.: Taking away training data by confidences. Applied Soft Computing  \textbf{199},  115320 (2026). \doi{10.1016/j.asoc.2026.115320}

\bibitem{nasr2019privacy}
Nasr, M., Shokri, R., Houmansadr, A.: Comprehensive privacy analysis of deep learning: Passive and active white-box inference attacks against centralized and federated learning. In: 2019 IEEE Symposium on Security and Privacy. pp. 739--753 (2019)

\bibitem{paszke2019pytorch}
Paszke, A., Gross, S., Massa, F., Lerer, A., Bradbury, J., Chanan, G., Killeen, T., Lin, Z., Gimelshein, N., Antiga, L., Desmaison, A., K{\"o}pf, A., Yang, E., DeVito, Z., Raison, M., Tejani, A., Chilamkurthy, S., Steiner, B., Fang, L., Bai, J., Chintala, S.: {PyTorch}: An imperative style, high-performance deep learning library. In: Advances in Neural Information Processing Systems. vol.~32, pp. 8026--8037 (2019)

\bibitem{shokri2017membership}
Shokri, R., Stronati, M., Song, C., Shmatikov, V.: Membership inference attacks against machine learning models. In: 2017 IEEE Symposium on Security and Privacy. pp. 3--18 (2017). \doi{10.1109/SP.2017.41}

\bibitem{song2017remember}
Song, C., Ristenpart, T., Shmatikov, V.: Machine learning models that remember too much. In: Proceedings of the 2017 ACM SIGSAC Conference on Computer and Communications Security. pp. 587--601 (2017)

\bibitem{sun2019backdoor}
Sun, Z., Kairouz, P., Suresh, A.T., McMahan, H.B.: Can you really backdoor federated learning? CoRR  \textbf{abs/1911.07963} (2019)

\bibitem{vero2023tableak}
Vero, M., Balunovic, M., Dimitrov, D.I., Vechev, M.: {TabLeak}: Tabular data leakage in federated learning. In: Proceedings of the 40th International Conference on Machine Learning. pp. 35051--35083 (2023)

\bibitem{wang2020attacktails}
Wang, H., Sreenivasan, K., Rajput, S., Vishwakarma, H., Agarwal, S., Sohn, J.y., Lee, K., Papailiopoulos, D.: Attack of the tails: Yes, you really can backdoor federated learning. In: Advances in Neural Information Processing Systems. vol.~33, pp. 16070--16084 (2020)

\bibitem{wei2020clientprivacy}
Wei, K., Li, J., Ding, M., Ma, C., Yang, H.H., Farokhi, F., Jin, S., Quek, T.Q.S., Poor, H.V.: A framework for evaluating client privacy leakages in federated learning. In: Computer Security -- ESORICS 2020. pp. 545--566 (2020)

\bibitem{wen2022fishing}
Wen, Y., Geiping, J., Fowl, L., Goldblum, M., Goldstein, T.: Fishing for user data in large-batch federated learning via gradient magnification. In: Proceedings of the 39th International Conference on Machine Learning. pp. 23668--23684 (2022)

\bibitem{xiao2017fashionmnist}
Xiao, H., Rasul, K., Vollgraf, R.: {Fashion-MNIST}: A novel image dataset for benchmarking machine learning algorithms. CoRR  \textbf{abs/1708.07747} (2017)

\bibitem{xie2020dba}
Xie, C., Huang, K., Chen, P.Y., Li, B.: {DBA}: Distributed backdoor attacks against federated learning. In: International Conference on Learning Representations (2020)

\bibitem{yang2019fml}
Yang, Q., Liu, Y., Chen, T., Tong, Y.: Federated machine learning: Concept and applications. ACM Transactions on Intelligent Systems and Technology  \textbf{10}(2),  1--19 (2019)

\bibitem{yang2022misclassification}
Yang, X., Luo, W., Zhang, L., Chen, Z., Wang, J.: Data leakage attack via backdoor misclassification triggers of deep learning models. In: 2022 4th International Conference on Data Intelligence and Security. pp. 61--66 (2022). \doi{10.1109/ICDIS55630.2022.00017}

\bibitem{yang2023imperceptible}
Yang, X., Luo, W., Zhou, Q., Chen, Z.: Training data leakage via imperceptible backdoor attack. In: 2023 IEEE Symposium Series on Computational Intelligence. pp. 1553--1559 (2023)

\bibitem{yin2021gradinversion}
Yin, H., Mallya, A., Vahdat, A., Alvarez, J.M., Kautz, J., Molchanov, P.: See through gradients: Image batch recovery via {GradInversion}. In: Proceedings of the IEEE/CVF Conference on Computer Vision and Pattern Recognition. pp. 16337--16346 (2021)

\bibitem{zhang2022neurotoxin}
Zhang, Z., Panda, A., Song, L., Yang, Y., Mahoney, M., Mittal, P., Kannan, R., Gonzalez, J.: Neurotoxin: Durable backdoors in federated learning. In: Proceedings of the 39th International Conference on Machine Learning. pp. 26429--26446 (2022)

\bibitem{zhao2020idlg}
Zhao, B., Mopuri, K.R., Bilen, H.: {iDLG}: Improved deep leakage from gradients. arXiv preprint arXiv:2001.02610  (2020)

\bibitem{zhu2019dlg}
Zhu, L., Liu, Z., Han, S.: Deep leakage from gradients. Advances in Neural Information Processing Systems  \textbf{32} (2019)

\end{thebibliography}

\end{document}